\documentclass[sigconf]{acmart}
\AtBeginDocument{%
  }

\setcopyright{acmlicensed}
\copyrightyear{2018}
\acmYear{2018}
\acmDOI{XXXXXXX.XXXXXXX}
\acmConference[]{}{}{}
\acmISBN{978-1-4503-XXXX-X/2018/06}

\usepackage{amssymb, amsmath, multirow, enumitem, colortbl, fancyhdr}
\definecolor{tabred}{rgb}{1, 0.7, 0.7}
\definecolor{taborange}{rgb}{1, 0.85, 0.7}
\definecolor{tabcapred}{rgb}{1, 0.65, 0.65}
\definecolor{tabcaporange}{rgb}{1, 0.7, 0.5}
\definecolor{figblue}{rgb}{0, 0.6902, 0.9412}
\definecolor{figorange}{rgb}{0.9294, 0.4902, 0.1921}
\newcommand{\best}{\cellcolor{tabred}}
\newcommand{\sbest}{\cellcolor{taborange}}

\renewcommand\footnotetextcopyrightpermission[1]{} 

\begin{document}

\fancypagestyle{standardpagestyle}{
  \fancyhf{}
  \renewcommand{\headrulewidth}{0pt}
  \renewcommand{\footrulewidth}{0pt}
}
\pagestyle{standardpagestyle} 

\title{Training-Free Hierarchical Scene Understanding for Gaussian Splatting with Superpoint Graphs}


\author{Shaohui Dai}
\authornote{Equal Contribution.}
\author{Yansong Qu}
\authornotemark[1]
\author{Zheyan Li} 
\author{Xinyang Li}

\author{Shengchuan Zhang}
\author{Liujuan Cao}
\authornote{Corresponding author.}
\email{daish@stu.xmu.edu.cn, caoliujuan@xmu.edu.cn}
\affiliation{%
\institution{Key Laboratory of Multimedia Trusted Perception and Efficient Computing, \\ Ministry of Education of China, Xiamen University, China}
}

\renewcommand{\shortauthors}{Dai et al.}

\begin{abstract}
Bridging natural language and 3D geometry is a crucial step toward flexible, language-driven scene understanding.
While recent advances in 3D Gaussian Splatting (3DGS) have enabled fast and high-quality scene reconstruction, research has also explored incorporating open-vocabulary understanding into 3DGS.
However, most existing methods require iterative optimization over per-view 2D semantic feature maps, which not only results in inefficiencies but also leads to inconsistent 3D semantics across views.
To address these limitations, we introduce a training-free framework that constructs a superpoint graph directly from Gaussian primitives.
The superpoint graph partitions the scene into spatially compact and semantically coherent regions, forming view-consistent 3D entities and providing a structured foundation for open-vocabulary understanding.
Based on the graph structure, we design an efficient reprojection strategy that lifts 2D semantic features onto the superpoints, avoiding costly multi-view iterative training.
The resulting representation ensures strong 3D semantic coherence and naturally supports hierarchical understanding, enabling both coarse- and fine-grained open-vocabulary perception within a unified semantic field.
Extensive experiments demonstrate that our method achieves state-of-the-art open-vocabulary segmentation performance, with semantic field reconstruction completed over 30× faster. Our code will be available at \url{https://github.com/Atrovast/THGS}.

\end{abstract}



\begin{teaserfigure}
\centering
  \includegraphics[width=\textwidth]{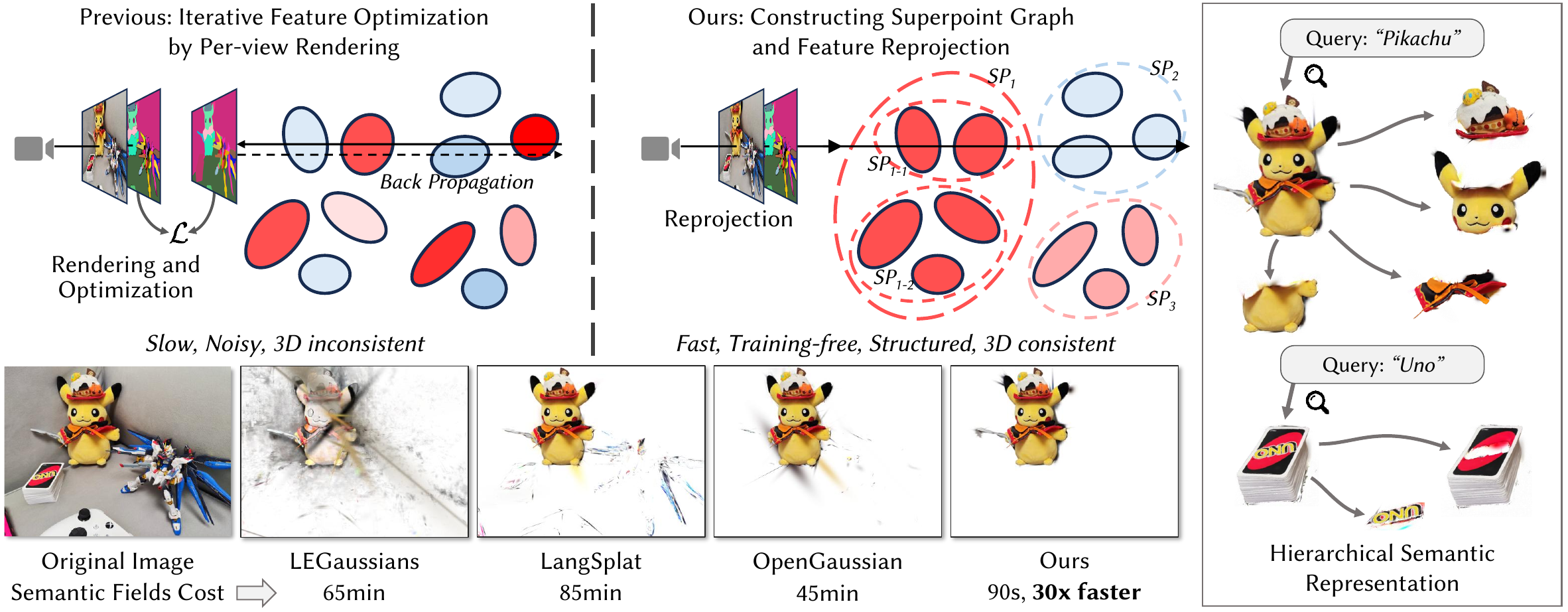}
  \caption{
We propose a method for open-vocabulary 3D scene understanding by integrating Gaussian Splatting with a superpoint graph. The top row shows that existing methods iteratively optimize features over Gaussian primitives, resulting in slow processing and inconsistent 3D semantics. In contrast, our approach clusters Gaussians into superpoints (denoted as \textit{SP}) and uses feature reprojection to build a semantic field. The bottom row highlights 3D consistency and speed improvements. The examples on the right demonstrate the capability of our work for open-vocabulary and hierarchical scene understanding.
}
  \label{fig:teaser}
\end{teaserfigure}

\maketitle

\section{Introduction}
In recent years, computer vision has made significant progress, particularly in developing systems that perceive and interact with the three-dimensional world.
One emerging challenge in this context is 3D open-vocabulary scene understanding, where machines are expected to identify and localize arbitrary regions within a 3D environment based on free-form natural language input.
This capability plays an important role in applications such as augmented reality and robotics, enabling users to refer to objects or regions in a scene using language descriptions rather than predefined labels.
As 3D perception continues to scale in both complexity and scene size, the demand for accurate, efficient, and semantically structured 3D understanding is becoming increasingly critical.

Given the scarcity of large-scale 3D datasets with language annotations, many approaches~\cite{kobayashi2022decomposing, tschernezki2022neural, lerf2023, huang2024nerf, liao2024ovnerf, qu2023sg} leverage vision-language models (VLMs), such as CLIP~\cite{radford2021clip} and LSeg~\cite{li2022lseg}, to distill open-vocabulary semantics into 3D scene representations.
These methods extract semantic features from 2D images and lift them into 3D, enabling language-driven interaction with reconstructed scenes, without requiring dense 3D supervision.
Among various 3D representations, 3D Gaussian Splatting (3DGS)~\cite{kerbl20233dgaussian} has recently emerged as a promising alternative for scene reconstruction due to its efficient optimization and real-time rendering. These advantages make it well-suited for embedding semantic information. Building on this, several recent works have explored combining image-derived language features with 3DGS to construct semantic fields under open-vocabulary settings \cite{lerf2023, shi2023legs, qin2023langsplat, zhou2023feature3dgs, zuo2024fmgs, wu2024opengaussian, liang2024supergseg}.

Despite recent progress, existing methods still suffer from two key limitations.
First, approaches like LangSplat~\cite{qin2023langsplat} and LEGaussians~\cite{shi2023legs} iteratively optimize semantic features on each view by rendering high-dimensional embeddings and backpropagation. However, they treat each Gaussian primitive in isolation and do not enforce consistency in 3D space, weakening semantic continuity across neighboring primitives. As a result, these methods often produce noisy semantic features and suffer from misalignment between 2D observations and the underlying 3D structure.
Second, semantic field reconstruction is often time-consuming, even surpassing the cost of appearance modeling. Semantic information is generally higher-level and lower-frequency than visual details and is expected to exhibit spatial smoothness in 3D. Yet, existing methods optimize dense high-dimensional embeddings across all primitives without spatial regularization, resulting in an ill-posed process that converges slowly.

To overcome the inefficiencies and semantic inconsistencies of existing methods, we rethink how semantic information should be represented and transferred in 3D, as illustrated in Figure~\ref{fig:teaser}. 
Instead of optimizing dense per-view embeddings for individual primitives, we observe that high-level semantics are typically low-frequency, spatially coherent, and better captured through structured abstraction.
This insight motivates our training-free framework, which builds hierarchical semantic fields by grouping Gaussian primitives into superpoints—compact clusters that are semantically homogeneous and geometrically coherent.
We begin with SAM-guided segmentation to partition the scene into superpoints that align well with object boundaries across views, thereby enforcing 3D semantic consistency from the start.
To further capture the multi-scale nature of real-world semantics, we progressively merge these superpoints into a multi-level graph, guided by segmentation masks of progressively coarser granularity. This hierarchy naturally supports open-vocabulary understanding at both object and part levels.
Rather than relying on slow, view-by-view optimization to distill semantics, we directly reproject 2D semantic features onto the superpoint graph using a simple and efficient aggregation mechanism.
This training-free, one-pass design eliminates costly iterative updates and enables semantic field reconstruction within minutes—achieving over 30× speedup compared to existing approaches.
Together, these innovations result in a scalable, hierarchical, and view-consistent solution for 3D open-vocabulary scene perception.

Our main contributions are summarized as follows:
\noindent
\begin{itemize}[leftmargin=10pt]
\item We propose a training-free framework for constructing 3D open-vocabulary semantic fields with strong spatial consistency and no iterative optimization.
\item A contrastive Gaussian partitioning strategy is incorporated to improve boundary precision and ensure semantic coherence.
\item We introduce a hierarchical merging and reprojection strategy that produces structured superpoint representations for multi-level semantic understanding.
\item Extensive experiments demonstrate that our method achieves state-of-the-art open-vocabulary scene understanding, while reducing semantic field reconstruction time by over 30×.
\end{itemize}

\section{Related Works}
\subsection{Gaussian Splatting}
3D Gaussian Splatting (3DGS) has emerged as a powerful and efficient representation for 3D scenes~\cite{kerbl20233dgaussian, Huang2DGS2024, li2024director3d, shen2025evolving, qu2025drag}. By modeling scenes as collections of explicit Gaussian primitives and rasterizing them for image synthesis, 3DGS achieves real-time rendering and fast reconstruction. 
Its point-based structure also provides flexibility for applications such as editing and manipulation. 
Subsequent works have aimed to improve 3DGS along several dimensions, including geometric fidelity, scalability, and robustness. To enhance surface accuracy, SuGaR~\cite{guedon2023sugar} introduces geometric constraints to Gaussian primitives. 2DGS~\cite{Huang2DGS2024} and PGSR~\cite{chen2024pgsr} replace 3D ellipsoids with 2D disks for tighter surface alignment. Scaffold-GS~\cite{scaffoldgs} proposes a hierarchical anchor-based structure to reduce redundancy and improve scalability. Other efforts focus on robustness under sparse-view conditions~\cite{li2024dngaussian, zhu2024fsgs}, anti-aliasing performance~\cite{yan2024aags, yu2024mipgsaa}, and adaptation to in-the-wild images~\cite{sabour2024spotlesssplats, wang2024we, wang2025look, kulhanek2024wildgaussians}.

\begin{figure*}[t]
    \centering
    \includegraphics[width=0.95\textwidth]{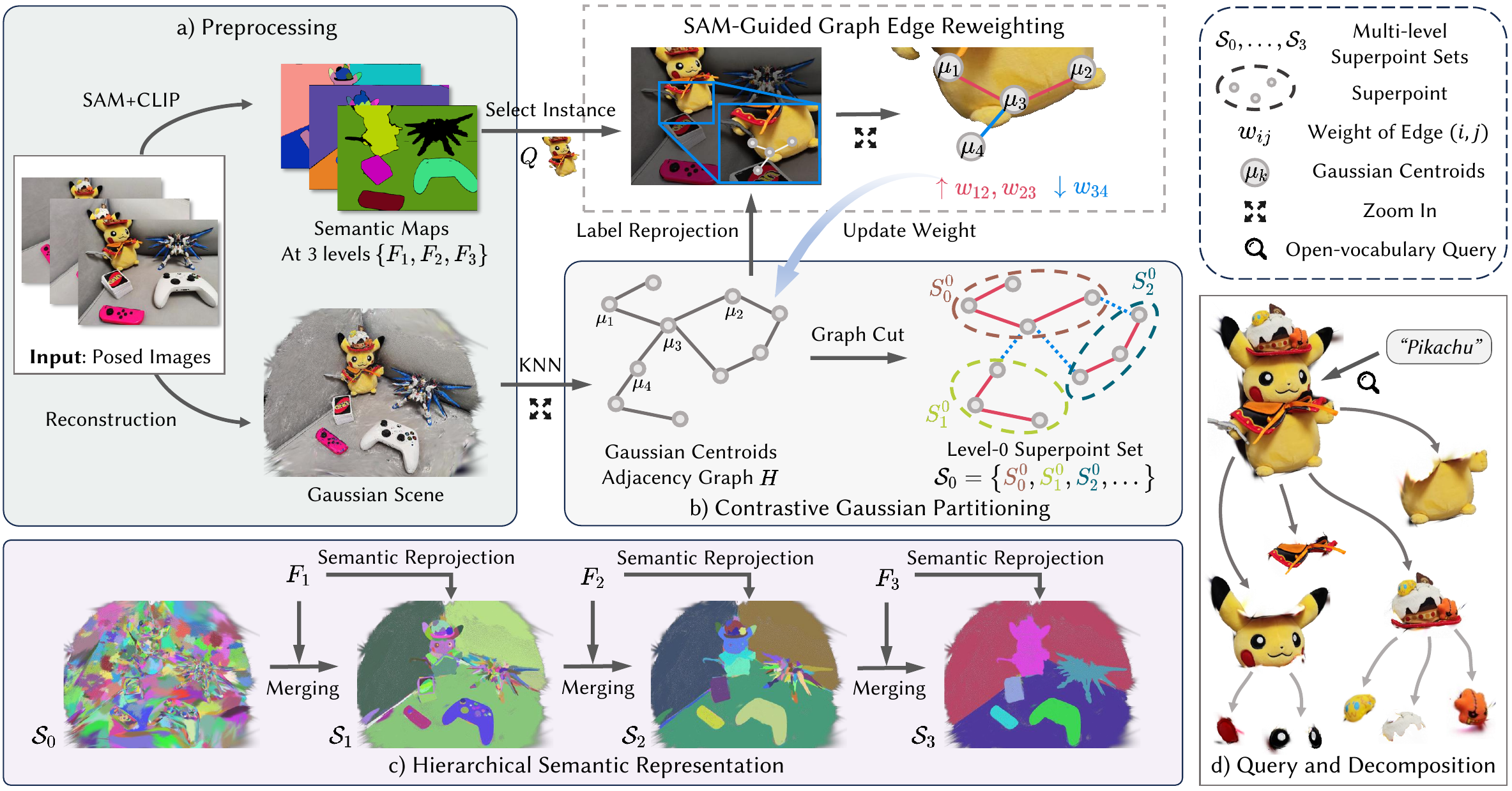}
    \caption{
Framework Overview.
a) Preprocessing: Scene reconstruction and extraction of 2D semantic maps.
b) Contrastive Gaussian Partitioning: A Gaussian adjacency graph is created, and its edge weights are adjusted using SAM-guided contrastive cues. The scene is then partitioned into superpoints.
c) Hierarchical Semantic Representation: Superpoints are progressively merged to form a multi-level superpoint graph, while semantic features are reprojected onto each level.
d) Query and Decomposition: The resulting hierarchical graph enables open-vocabulary query and part-based decomposition of scene objects.
    }
    \label{fig:pipeline}
\end{figure*}

\subsection{Open-vocabulary Scene Understanding with Radiance Fields}
Vision-language models (VLMs) such as CLIP~\cite{radford2021clip} have enabled open-vocabulary reasoning in visual domains, inspiring a series of methods for 3D semantic understanding. Early works~\cite{kobayashi2022decomposing, liu20233dovs, tschernezki2022neural, lerf2023} lift CLIP features from 2D views into NeRF scenes, allowing text-based localization. However, the implicit volumetric nature of NeRF limits their training efficiency and real-time utility. 
Recent methods based on Gaussian Splatting offer faster reconstruction and explicit point-based representations. LangSplat~\cite{qin2023langsplat} compresses CLIP features using an autoencoder and incorporates SAM masks for hierarchical supervision. LEGaussians~\cite{shi2023legs} proposes a quantized and smoothed embedding strategy to preserve rendering quality. Gaussian Grouping~\cite{ye2023gaussiangroup} uses SAM~\cite{kirillov2023sam} priors to improve boundary accuracy and region separation. 
GOI~\cite{qu2024goi} introduces an optimizable semantic-space Hyperplane to achieve more accurate open-vocabulary semantic perception at test time. OpenGaussian~\cite{wu2024opengaussian} applies contrastive learning with SAM supervision to construct object-level semantic fields. SuperGSeg~\cite{liang2024supergseg} shares a similar motivation to ours, drawing from superpoint-based methods. It performs multi-view instance and hierarchical segmentation on Scaffold-GS by leveraging Super-Gaussians to enhance interpretability.

\subsection{Superpoint-based Point Cloud Segmentation}
Superpoint-based methods have gained increasing attention in 3D point cloud segmentation due to their efficiency and ability to preserve local geometric structure \cite{landrieu2018spg, robert2023spt, landrieu2019sp1, hui2021sp2, yin2024sai3d, nguyen2024open3dis}. Inspired by superpixels in 2D image segmentation \cite{achanta2012slic}, these approaches group neighboring points into compact and geometrically consistent regions called superpoints, which serve as mid-level representations for downstream semantic reasoning.
SPG~\cite{landrieu2018spg} and SPT~\cite{robert2023spt} propose building superpoint graphs to model spatial and geometric relations between regions, and further refine superpoint features using neural networks. SAI3D~\cite{yin2024sai3d} leverages SAM to iteratively merge superpoints into segments aligned with multi-view 2D masks, producing consistent instance-level groupings without requiring semantic supervision. Open3DIS~\cite{nguyen2024open3dis} combines superpoints with a 3D instance segmentation network and embeds CLIP features at the instance level to support open-vocabulary segmentation.
These approaches demonstrate the effectiveness of superpoint for scalable and structured 3D semantic understanding.

\section{Methods}
Given a set of posed images, we first reconstruct a 3D Gaussian scene $G$ using Gaussian Splatting~\cite{kerbl20233dgaussian}. Our goal is to extend $G$ with open-vocabulary semantics, enabling users to query and segment objects in both 2D and 3D using natural language descriptions.

Unlike previous methods that rely on per-view optimization and lack mechanisms for global consistency, we construct a unified 3D semantic representation directly on a superpoint graph built from Gaussian primitives (GPs). This design enables fast, training-free reconstruction of a view-consistent semantic field, as illustrated in Figure~\ref{fig:pipeline}.
We begin by preprocessing each image with frozen vision-language models to extract 2D feature maps and perform scene reconstruction. We treat the Gaussian primitives in $G$ as a point cloud and perform a contrastive Gaussian partitioning to segment the scene into simple, semantically homogeneous clusters, referred to as \textit{superpoints}. For the over-segmented superpoints, we adopt a hierarchical merging strategy that groups adjacent superpoints into semantically distinct larger clusters, resulting in a multi-level \textit{superpoint graph} aligned with the levels of SAM masks. During this process, semantic features are reprojected onto the multi-level superpoints, enabling efficient construction of a semantic field.

\subsection{Preliminaries: Gaussian Splatting}
\label{subsec_3dgs}
Gaussian Splatting represents a 3D scene as a collection of Gaussian primitives, which extend traditional point clouds with additional shape, opacity, and appearance parameters. Given known camera poses, the scene is rendered by splatting these primitives onto the image plane and rasterizing the resulting 2D ellipses in a differentiable manner.

Each Gaussian primitive is parameterized by a centroid $\mu \in \mathbb{R}^3$, a 3D covariance matrix $\Sigma$, opacity $\alpha$, and spherical harmonics coefficients $c$ for view-dependent color. To ensure that $\Sigma$ remains valid and interpretable, it is factorized as:
\begin{equation}
\Sigma = R S S^T R^T,
\end{equation}
where $R$ is a rotation matrix and $S$ is a scaling matrix. The full set of learnable parameters for the $i$-th Gaussian primitive is given by:
$\theta_i = \{\mu_i, c_i, \alpha_i, R_i, S_i\}$.

Rendering is performed via volumetric alpha compositing. For each pixel, the color $C$ is computed as:
\begin{equation}
\label{eq:render}
C = \sum_{i \in G'} c_i \alpha_i T_i,
\end{equation}
where $G'$ is the depth-sorted set of Gaussian primitives contributing to the pixel, and $T_i = \prod_{j=1}^{i-1}(1 - \alpha_j)$ denotes the accumulated transmittance up to the $i$-th primitive.

To achieve high-fidelity surface reconstruction, we adopt 2D Gaussian Splatting (2DGS)~\cite{Huang2DGS2024}, which improves geometric accuracy by modeling each surface element as a 2D disk rather than a 3D ellipsoid. In addition to higher geometric precision, 2DGS provides surface normals $\mathbf{n}_i$ for each primitive, which can be utilized in the following processing.

\subsection{Contrastive Gaussian Partitioning}
\label{sec:partition}
We begin by partitioning the Gaussian primitives in the scene into a set of spatially compact and semantically coherent clusters, referred to as \textit{superpoints}. 
To obtain these superpoints, we treat the GPs as a 3D point cloud and construct an adjacency graph based on spatial proximity. A graph partitioning algorithm is then applied to segment this graph into geometrically simple clusters.

However, unlike traditional point-based partitioning, GPs possess spatial extent and may span multiple semantic regions when projected onto the image plane. As a result, naive clustering can group semantically inconsistent content into a single superpoint. 
To mitigate semantic ambiguity, we use SAM-generated 2D masks as guidance to reweight the graph edges, encouraging connections within the same region and suppressing those across boundaries.
This reweighting introduces a contrastive constraint into the partitioning process, encouraging the formation of semantically homogeneous superpoints aligned with object boundaries and consistent across views.

\subsubsection{Graph-based Gaussian Partitioning}
To adapt point cloud segmentation techniques to Gaussian primitives, we ignore the volumetric nature of each GP and represent it solely by its geometric centroid $\mu$. The set of centroids is denoted as $G_\mu$. We construct an undirected $K$-nearest neighbor graph by connecting each GP to its $K$ closest neighbors in Euclidean space. The resulting adjacency graph $H$ is defined as the combination of the vertex set $G_\mu$ and the edge set $E$:
\begin{equation}
    H = (G_\mu, E), \quad E = \{(i, j) \mid \mu_i, \mu_j \in G_\mu \}.
\end{equation}

For each node $\mu_i$, we construct a feature vector $p_i$ by concatenating its position and additional GP features, such as color or normal:
\begin{equation}
    p_i = \operatorname{concat}\{\mu_i, c_i, \mathbf n_i, \ldots \}, \quad \mu_i \in G_\mu.
\end{equation}
And the edge weight $w_{ij}$ is computed inversely proportional to the normalized feature distance between $p_i$ and $p_j$:
\begin{equation}
    w_{ij} = \frac{1}{1 + \operatorname{dis}(\mu_i, \mu_j)}, \ \operatorname{dis}(\mu_i, \mu_j) = \frac{\|p_i - p_j\|_2}{\frac{1}{|E|}  \sum_{(k, l) \in E} \|p_k - p_l\|_2}.
\end{equation}
This feature construction approach is adapted from~\cite{landrieu2018spg, robert2023spt}, with minor modifications tailored to Gaussian primitives.

With the graph weights defined, we apply the Cut Pursuit algorithm~\cite{landrieu2017cutp} to partition the graph into superpoints. Edges with lower weights are more likely to be cut, encouraging the separation of geometrically or semantically dissimilar regions.
The output of the algorithm yields a set of constant connected components in the graph, which we define as level-$0$ superpoint set $\mathcal S_0$.

\subsubsection{SAM-guided Graph Edge Reweighting}

To encourage semantic consistency during superpoint partitioning, we refine the edge weights in the adjacency graph using SAM-predicted segmentation masks before applying Cut Pursuit. While the initial graph captures geometric proximity, it does not account for semantic alignment. Since Gaussian primitives have non-zero spatial extent, they may overlap multiple regions or objects, leading to semantic ambiguity.

To mitigate this, we introduce a mask-guided edge reweighting strategy. For each view, SAM masks are indexed with integer labels and reprojected onto the GPs (see Sec.~\ref{sec:feat_proj}). For any connected pair of nodes $\mu_i$ and $\mu_j$, if their mask labels differ ($l_i \ne l_j$), the connection is down-weighted; otherwise, it is strengthened. 

We update the edge weights $w_{ij}$ in the adjacency graph accordingly:
\begin{equation}
w_{ij}' = w_{ij} + \delta_+ \cdot [l_i = l_j] - \delta_- \cdot [l_i \neq l_j],
\end{equation}
where $[\cdot]$ is the Iverson bracket, $\delta_+$ and $\delta_-$ control the influence of SAM masks on the graph structure. 
In practice, we ensure that $w_{ij}'>0$ to preserve meaningful graph connectivity.
To account for reduced reliability in distant projections~\cite{yin2024sai3d}, $\delta_+$ and $\delta_-$ are scaled by a depth-aware decay function, diminishing the effect of masks when the corresponding GPs are far from the camera.

The updated graph is then used in the Cut Pursuit algorithm to generate superpoints. This reweighting step helps align superpoint boundaries more closely with semantic object boundaries, promoting intra-region consistency and inter-region separation.

\subsection{Feature Reprojection for Semantic Guidance}
\label{sec:feat_proj}
Our framework requires transferring 2D segmentation information from SAM masks to 3D Gaussian primitives to support key components such as edge reweighting, hierarchical merging, and semantic assignment. We propose a training-free feature reprojection mechanism that robustly associates 2D mask labels with 3D Gaussian primitives in the presence of boundary ambiguity.

This process consists of two steps: (1) encoding each SAM mask label into a latent semantic embedding, and (2) reprojecting these embeddings onto Gaussian primitives through a rendering-guided aggregation scheme. This enables direct and reliable assignment of semantic features from 2D to 3D without iterative optimization.

\subsubsection{Latent label encoding.}

For each SAM-generated mask $m_t$, where $t \in \{1, 2, ..., M\}$, we assign a unique integer label $t$ to identify the mask. However, directly reprojecting raw integer labels onto 3D Gaussian primitives often leads to semantic ambiguity, especially near region boundaries. One-hot encodings are also suboptimal due to their variable length and high dimensionality.

To overcome these limitations, we map each label $t$ to a fixed-dimensional latent vector $Y_t \in \mathbb{R}^D$, sampled from a standard normal distribution and normalized to unit length:
\begin{equation} 
    Y_t = \operatorname{normalize}(\operatorname{rand}(D)), \quad t \in \{1, 2, ..., M\}.
\end{equation}
Here, $M$ is the total number of SAM-predicted masks.
Inspired by spline positional encoding~\cite{wang2021splinepe}, this approach distributes the label embeddings uniformly on the hypersphere, making them maximally discriminative under cosine similarity. In practice, we assign the latent label embedding $Y_t$ to every pixel belonging to mask $m_t$, resulting in a pixel-aligned semantic feature map $\widetilde{Y}$ in image space.

\subsubsection{Rendering-guided reprojection.}

For each view, we rasterize the 3D scene using Gaussian Splatting and compute the transmittance $T_{k}^{\mathbf{x}}$ of each Gaussian primitive $k$ along the pixel ray $\mathbf{x}$, following the volumetric rendering formulation (Equation~\ref{eq:render}). Since transmittance reflects the contribution of each GP to the final pixel color, we use it as a soft weighting factor to aggregate 2D semantic information onto the GPs.

The latent feature $y_k$ of Gaussian primitive $k$ is computed by aggregating label embeddings $\widetilde{Y}^{\mathbf{x}}$ from all contributing pixels, weighted by their transmittance values $T_k^{\mathbf{x}}$:
\vspace{-2pt}
\begin{equation}
    y_k = \frac{\sum_{\mathbf{x}} T_{k}^{\mathbf{x}} \widetilde{Y}^{\mathbf{x}}}{\sum_{\mathbf{x}} T_{k}^{\mathbf{x}}}.
\end{equation}

We then assign a discrete semantic label $l_k$ by matching latent feature $y_k$ to the closest entry $Y_t$ in the label embedding set $\{Y_t\}_{t=1}^M$ via cosine similarity:
\begin{equation}
    l_k = \operatorname*{argmax}_{t \in \{1,2,...,M\}} \cos\left ( y_k, Y_t\right ).
\end{equation}

This feature reprojection mechanism effectively preserves semantic boundaries and reduces ambiguity near object edges.

\subsection{Hierarchical Semantic Representation}
To build structured and interpretable scene semantics beyond the over-segmented superpoints $\mathcal{S}_0$ (from Sec.~\ref{sec:partition}), we adopt a bottom-up hierarchical merging strategy inspired by SAI3D~\cite{yin2024sai3d}. While SAI3D is initially developed for point-level instance grouping, we extend this approach to operate on Gaussian primitives and generalize it to construct a multi-level representation.

We use multi-level SAM masks $\{F_1, F_2, F_3\}$ to guide hierarchical merging. Each corresponds to a specific level of segmentation granularity. At each level, adjacent superpoints are aggregated based on their affinity score, producing progressively larger clusters. This process yields a multi-level superpoint graph $\mathcal{S}_1$, $\mathcal{S}_2$, and $\mathcal{S}_3$, which represent sub-parts, parts, and complete objects, respectively. Each level builds upon the previous one.

Semantic features are assigned to superpoints at all levels via the feature reprojection scheme, resulting in a hierarchical semantic field. This field enables open-vocabulary understanding at multiple levels and supports object decomposition.

\subsubsection{Superpoint Merging based on Affinity Score}
We adopt a hierarchical superpoint merging strategy inspired by SAI3D~\cite{yin2024sai3d}, which merges superpoints based on their affinity scores. The affinity scores reflect the likelihood that two regions belong to the same object instance. We compute the score using 2D SAM-generated masks as guidance.

At level $q$, for each pair of superpoints $S^q_u$ and $S^q_v$, we reproject the 2D masks $m_t \ (t \in \{1,2,... ,M\})$ onto the scene using the rendering-guided method described in Sec.~\ref{sec:feat_proj}.
For $S^q_u$ and $S^q_v$, we compute a histogram feature $\mathbf{h}_u^q$ and $\mathbf{h}_v^q$ that captures the distribution of its constituent Gaussian primitives across the reprojected masks.

The affinity score between $S^q_u$ and $S^q_v$ is then defined as the cosine similarity between their histogram features:
\begin{equation}
A_{u, v}^q = \cos\left( \mathbf{h}_u^q, \mathbf{h}_v^q \right).
\end{equation}

The affinity scores are averaged across views. 
Adjacent superpoints with affinity scores exceeding a predefined threshold are merged to form coarser superpoints.

\subsubsection{Semantic Feature Assignment}

The resulting multi-level superpoints are spatially coherent and significantly sparser than the original Gaussian primitives (typically by two orders of magnitude), making semantic field construction more efficient. Moreover, since both superpoint partitioning and hierarchical merging are guided by SAM masks, the resulting superpoints exhibit strong alignment with object boundaries. This allows us to directly reproject semantic features to superpoints without any further optimization.

During the merging process, we establish the correspondence between each superpoint and the set of masks it overlaps with. We then assign a semantic feature to each superpoint by computing a weighted average of the features from the corresponding masks. Let $f_t$ denote the semantic feature of mask $m_t$. The semantic feature $f(S_k^q)$ of superpoint $S_k^q$ is computed as:
\vspace{-2pt}
\begin{equation} 
f(S_k^q) = \sum_{t=1}^{M} \omega_t \cdot f_t, \quad \omega_t = \frac{\operatorname{NumVis}(S_k^q, m_t)}{\left |S_k^q \right |}. 
\end{equation}
Here, $\operatorname{NumVis}(S_k^q, m_t)$ denotes the number of visible Gaussian primitives in $S_k^q$ that are assigned to mask $m_t$, and $\left |S_k^q \right |$ is the total number of GPs in the superpoint. The weight $\omega_t$ thus reflects the proportion of $S_k^q$ associated with mask $m_t$. We then aggregate multi-view features to obtain overall superpoint semantic features.

Semantic features derived from SAM masks of three levels are reprojected onto the corresponding superpoint levels $\mathcal{S}_1$, $\mathcal{S}_2$, and $\mathcal{S}_3$, representing sub-parts, parts, and whole objects respectively. These layers are nested within a hierarchical superpoint graph, forming a unified semantic field that supports multi-level, open-vocabulary scene understanding.

\begin{figure*}[ht]
    \centering
    \includegraphics[width=0.95\textwidth]{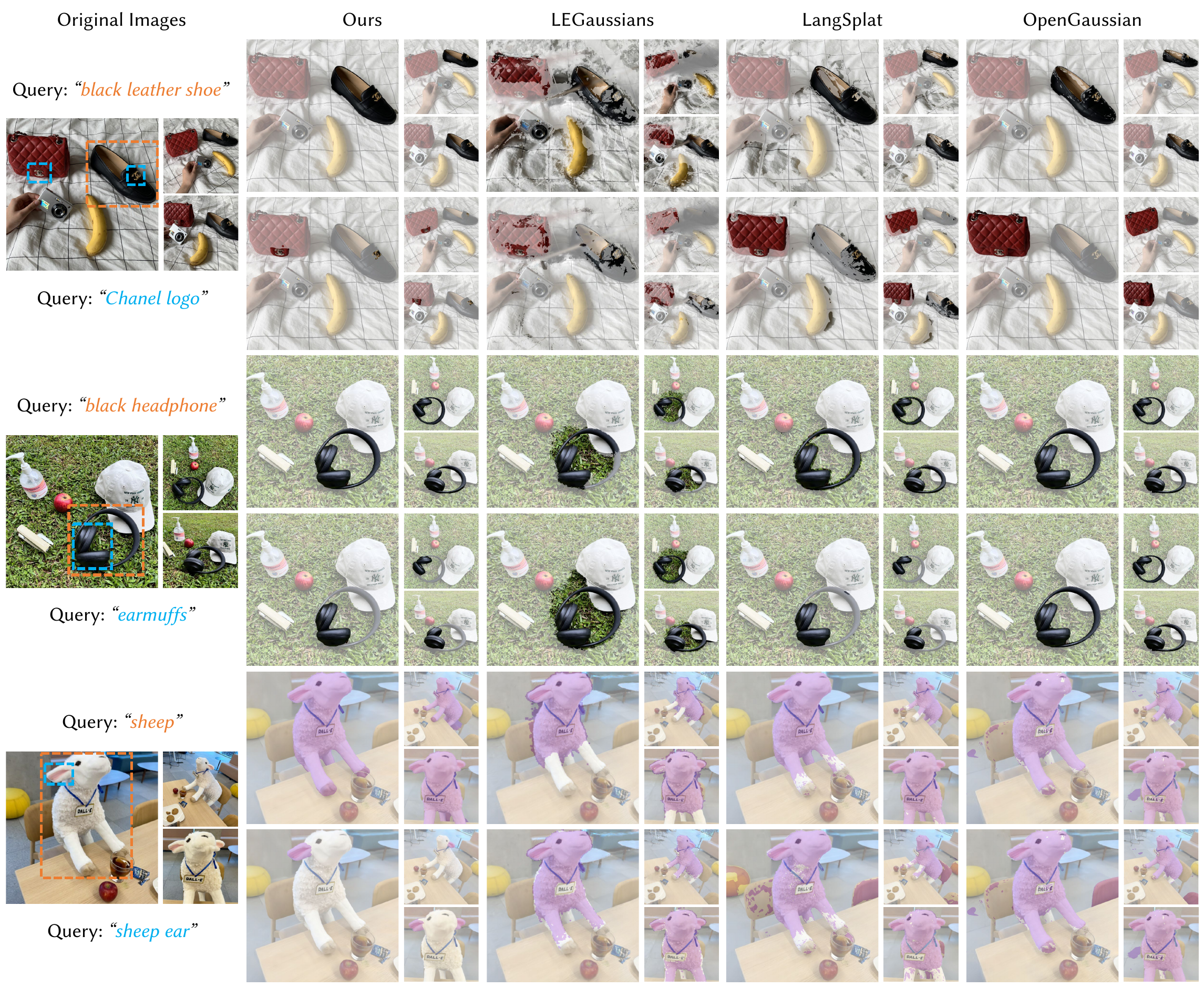}
    \caption{
    Qualitative comparisons of open-vocabulary segmentation on multi-view 2D images.
We compare our method with LEGaussians~\cite{shi2023legs}, LangSplat~\cite{qin2023langsplat}, and OpenGaussian~\cite{wu2024opengaussian}. Each scene includes an object- and a part-level query. Query results are highlighted, and corresponding ground-truth regions are marked with \textcolor{figorange}{orange} (object) and \textcolor{figblue}{blue} (part) bounding boxes.
}
    \label{fig:expt_2d}
\end{figure*}

\subsection{Evaluation}
\label{sec:eval}

We follow the evaluation protocol proposed in LERF~\cite{lerf2023}, adapted to our superpoint-based hierarchical representation. For each text query, we compute a relevance score between the query embedding $\phi_{\text{qry}}$ and each superpoint semantic feature $\phi_{\text{sp}}$ across one or multiple hierarchy levels.

For each superpoint, the relevance score is defined as:
\begin{equation}
\min_i \frac{\exp(\phi_{\text{sp}} \cdot \phi_{\text{qry}})}{\exp(\phi_{\text{sp}} \cdot \phi_{\text{qry}}) + \exp(\phi_{\text{sp}} \cdot \phi_{\text{canon}}^i)},
\end{equation}
where $\phi_{\text{canon}}^i$ denotes the CLIP embedding of a predefined canonical concept (i.e., ``object'', ``things'', ``stuff'', ``texture''). This contrastive scoring helps disambiguate the target object from generic or background categories.

After computing relevance scores, we select the top-ranked superpoints as query results, and retrieve their associated Gaussian primitives for further rendering or analysis. The query can be performed at a single hierarchical level or jointly across multiple levels for more robust localization.

To visualize the result in 2D, we do not render the full high-dimensional semantic field. Instead, we rasterize a binary presence mask by assigning a value of $b_k = 1$ to each selected GP and rendering it using standard volumetric accumulation:
\begin{equation}
B = \sum_{k \in G'} b_k \alpha_k T_k .
\end{equation}
The resulting soft mask $B$ is thresholded at 0.5 to obtain a binary segmentation map.

\begin{table*}[t]
  \caption{Quantitative comparison of 2D open-vocabulary segmentation on the LERF-OVS~\cite{lerf2023} and 3DOVS~\cite{liu20233dovs} datasets. ``SR Time'' refers to the Semantic Field Reconstruction Time. The best and second-best results are highlighted in \textcolor{tabcapred}{red} and \textcolor{tabcaporange}{orange}.}
  \label{tab:lerf_3dovs}
   {\begin{tabular}{c |cccc c|c |ccccc c|c}
    \toprule
    \multirow{2}{*}{Methods} & \multicolumn{5}{c|}{LERF-OVS mIoU (\%)} & LERF & \multicolumn{6}{c|}{3DOVS mIoU (\%)} & \makebox[0.05\linewidth][c]{3DOVS} \\
     & \makebox[0.06\linewidth][c]{Figurines} & \makebox[0.047\linewidth][c]{Ramen} & \makebox[0.047\linewidth][c]{Teatime} & \makebox[0.047\linewidth][c]{Waldo} & \textbf{Overall} & \makebox[0.05\linewidth][c]{SR Time} 
     & \makebox[0.039\linewidth][c]{Bed} & \makebox[0.039\linewidth][c]{Bench} & \makebox[0.039\linewidth][c]{Lawn} & \makebox[0.039\linewidth][c]{Room} & \makebox[0.039\linewidth][c]{Sofa} &\textbf{Overall}& \makebox[0.05\linewidth][c]{SR Time}\\
    \midrule
    LEGau. \cite{lerf2023} & 27.60    & 13.75    &45.21  & 23.71  & 27.57 &65min  
    &11.21 &62.66 &58.61 &64.03 &9.02 &41.11  & 75min\\    
    
    LangSplat \cite{lerf2023} & 44.7    & \best 51.2    & 65.1  & 44.5 & \sbest 51.4   & 85min    
    &92.5 & \sbest 94.2 &96.1 & \best 94.1 & \sbest 90.0 & \sbest 93.4  & 90min\\    
    GOI \cite{qu2024goi} & 36.89 & 35.09 & \sbest 66.56 & \sbest 45.23 & 45.94    & \sbest 15min 
     &\best 97.70 & 73.24 & \sbest 96.26 & 73.91 & 85.16 & 85.26  & \sbest 14min \\
     
    OpenGau. \cite{wu2024opengaussian} & \best 69.73  & 21.11    & 63.35  & 34.91 & 47.28  & 45min  
    & 57.87 & 74.94 & 63.95 & 40.16 & 70.56 & 61.49   & 55min\\   
    
    Ours & \sbest 57.30 & \sbest 43.46  & \best 68.33 &\best 50.65   & \best 54.94 & \best 90s 
    & \sbest 96.01 & \best 95.14 & \best 96.88 & \sbest 92.71 & \best 93.89 & \best 94.93 & \best 25s \\   

  \bottomrule
\end{tabular}}
\end{table*}

\section{Implementation Details}

Our method is implemented on top of 2D Gaussian Splatting~\cite{Huang2DGS2024}. For semantic feature extraction, we follow LangSplat~\cite{qin2023langsplat}, using OpenCLIP ViT-B/16 as the vision-language encoder, alongside SAM ViT-H for 2D segmentation guidance. To support multi-level representation, we generate SAM masks at three different levels and obtain a CLIP embedding for each mask by feeding the masked region into the CLIP image encoder.

Our hierarchical semantic field construction is fully training-free and takes less than 2 minutes on average using a single RTX 3090 GPU. The runtime may vary depending on the number of input views and scene complexity.

\section{Experiments}
\subsection{Experimental Setup}

We evaluate our method on three datasets: LERF-OVS, 3DOVS, and ScanNet, covering both open-vocabulary 2D segmentation and 3D semantic understanding.

\textbf{Datasets.}  
The LERF-OVS benchmark is derived from the LERF dataset~\cite{lerf2023}, which consists of 14 real-world indoor scenes with posed multi-view RGB images. LangSplat~\cite{qin2023langsplat} selects four scenes (\textit{figurines}, \textit{ramen}, \textit{teatime}, and \textit{waldo kitchen}) from LERF and adds 2D mask annotations along with natural language descriptions for selected objects.
3DOVS dataset~\cite{liu20233dovs} includes 10 indoor and outdoor scenes featuring long-tail objects captured under diverse poses and backgrounds. Each scene provides 2D segmentation masks and text descriptions. We evaluate on five commonly used scenes: \textit{bed}, \textit{bench}, \textit{lawn}, \textit{sofa}, and \textit{room}. ScanNet~\cite{dai2017scannet} is a large-scale RGB-D dataset with over 1,500 indoor scenes. It offers RGB-D sequences, point clouds from scans, and semantic annotations. Following OpenGaussian~\cite{wu2024opengaussian}, we evaluate on 10 random scenes.

\textbf{Evaluation Protocol.}  
For all datasets, we reconstruct a 3D semantic field from posed multi-view images. Given a natural language query, the relevant region is retrieved via computing the relevance between text query and superpoint features, as described in Sec.~\ref{sec:eval}. 
For LERF-OVS and 3DOVS, we reproject the retrieved 3D semantic regions onto 2D views and compare them with ground-truth masks. We report per-scene and overall mean Intersection-over-Union (mIoU), along with the average semantic field reconstruction time. 
For ScanNet, we evaluate directly in 3D by first assigning predicted instance-level labels to Gaussian primitives and then comparing them with the ground truth point-level semantic annotations.
We report mean Intersection-over-Union (mIoU) and mean class accuracy (mAcc), which are 
computed over 19, 15, and 10-class subsets following the evaluation of OpenGaussian~\cite{wu2024opengaussian}.

\begin{table}[h]
\caption{Quantitative comparison for 3D semantic segmentation on ScanNet~\cite{dai2017scannet} dataset.}
\centering
\resizebox{\linewidth}{!}
{

\begin{tabular}{c|cccccc}
\toprule
\multirow{2}{*}{Methods} & \multicolumn{2}{c}{19 classes} & \multicolumn{2}{c}{15 classes} & \multicolumn{2}{c}{10 classes}  \\
                     & mIoU     & mAcc & mIoU  & mAcc & mIoU  & mAcc \\ 
\midrule

LangSplat~\cite{qin2023langsplat}    & 3.78       & 9.11       & 5.35       & 13.20       & 8.40       & 22.06       \\
LEGau.~\cite{shi2023legs}  & 3.84       & 10.87       & 9.01       & 22.22       & 12.82       & 28.62       \\
OpenGau.~\cite{wu2024opengaussian} & \sbest {24.73} & \sbest{41.54} & \sbest{30.13} & \sbest {48.25} & \sbest{38.29} & \sbest{55.19} \\ 
Ours & \best 34.39 & \best 50.74 
& \best 39.61 & \best 57.07 
& \best 46.38 & \best 64.74 \\ 
\bottomrule
\end{tabular}
}
\label{tab:scannet}
\end{table}

\begin{figure*}[ht]
    \centering
    \includegraphics[width=0.96\textwidth]{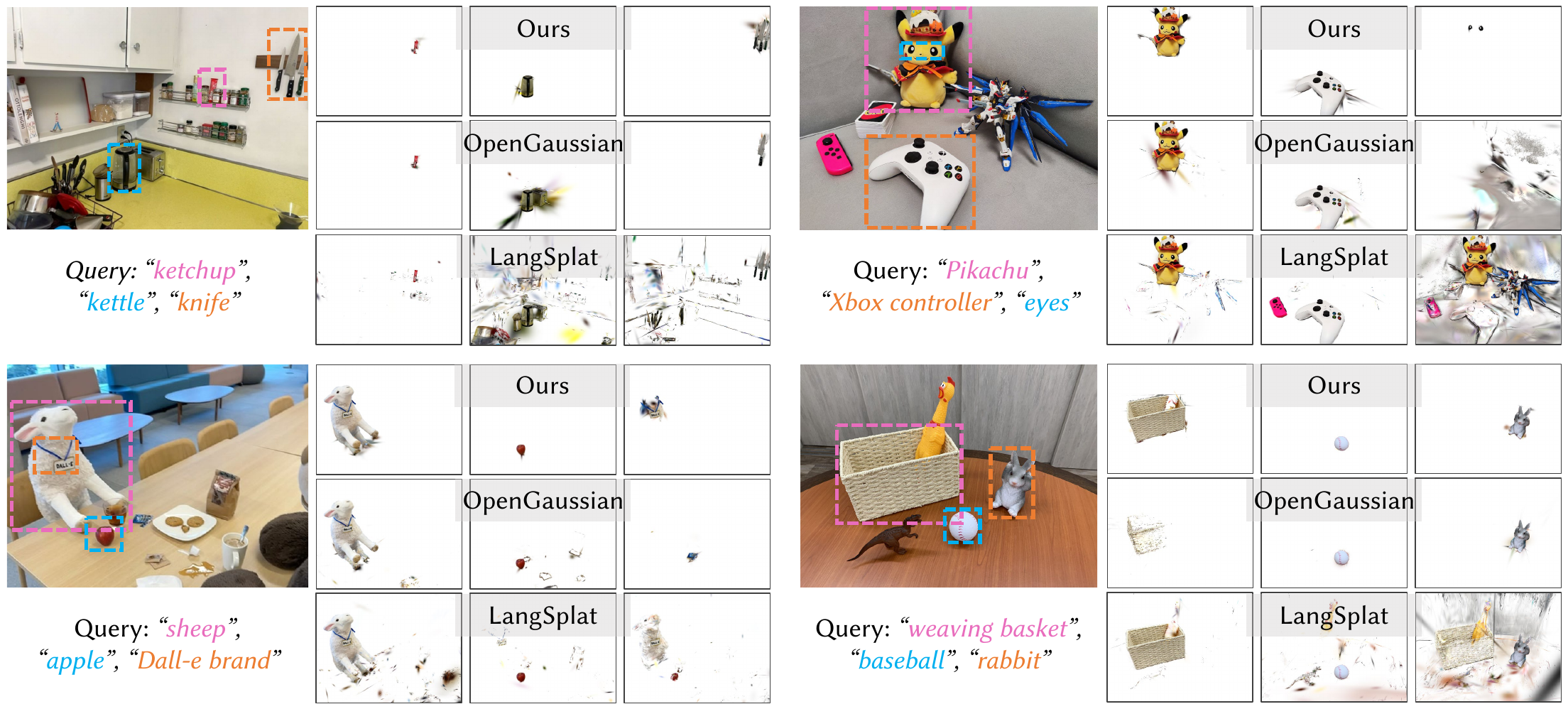}
    \caption{
    Qualitative comparison of open-vocabulary 3D segmentation.
We compare our method with OpenGaussian~\cite{wu2024opengaussian} and LangSplat~\cite{qin2023langsplat} by visualizing the predicted 3D Gaussian primitives. 
The queried regions are annotated with colored bounding boxes on the original images to indicate object locations.
}
    \label{fig:expt_3d}
\end{figure*}

\subsection{Comparisons}
We conduct a comparative evaluation of our approach in contrast with LEGaussians~\cite{shi2023legs}, LangSplat~\cite{qin2023langsplat}, GOI~\cite{qu2024goi}, and OpenGaussian~\cite{wu2024opengaussian}.

\textbf{Qualitative Results.}
We present the qualitative results generated by our method, along with comparisons to other approaches. Figures~\ref{fig:expt_2d} and~\ref{fig:expt_3d} provide a detailed overview of open-vocabulary scene understanding performance on the LERF-OVS and 3DOVS datasets, evaluated at both the 2D pixel level and the 3D Gaussian primitive level.

Figure~\ref{fig:expt_2d} presents the multi-view segmentation results on 2D images. For each scene, we provide two natural language queries—at object and part levels—to demonstrate both cross-view consistency and hierarchical understanding.
LEGaussians, trained only with 2D supervision and lacking geometric priors, often produces blurry and imprecise boundaries.
Our method and LangSplat, both guided by SAM, show multi-level understanding and can distinguish object and its part (e.g., ``headphones'' vs. ``earmuffs''). However, LangSplat struggles with finer categories like ``sheep ear'' and shows inconsistency in cases like the ``Chanel logo'', where our method remains accurate.
OpenGaussian and our approach both retrieve in 3D and project to 2D, enabling consistent multi-view appearance. Yet, OpenGaussian’s fixed-size codebook limits its part-level expressiveness, though it performs well at coarse object localization.

Figure~\ref{fig:expt_3d} shows qualitative results rendered directly from querying on 3D Gaussian primitives. We compare our method against two representative baselines. 
LangSplat, which relies on per-view 2D optimization, produces spatially inconsistent and noisy 3D segmentations, with many scattered or misaligned points.
Both our method and OpenGaussian yield relatively clean 3D segmentation results. However, OpenGaussian often lacks spatial coherence, mistakenly including scattered points from unrelated objects (e.g., ``sheep'' and ``apple'' in the bottom left scene).
In contrast, our superpoint-based framework enforces structural regularity and spatial compactness, resulting in more coherent and reliable segmentations.

Overall, our method offers more coherent 3D segmentation, better view consistency, and supports multi-level perception in both 2D and 3D.

\textbf{Quantitative Results.}
Tables \ref{tab:lerf_3dovs} and \ref{tab:scannet} summarize the quantitative results across three benchmarks. Our method achieves state-of-the-art performance on both 2D and 3D open-vocabulary segmentation tasks.
On LERF-OVS and 3DOVS (Table~\ref{tab:lerf_3dovs}), we evaluate open-vocabulary on 2D images. Our method achieves strong per-scene performance and outperforms existing methods in overall mIoU. 
On ScanNet (Table~\ref{tab:scannet}), we directly evaluate in 3D against ground-truth semantic labels. Our method surpasses LangSplat and LEGaussian by a large margin and improves over OpenGaussian by approximately 9\% in both mIoU and mAcc.

In addition to segmentation accuracy, Table~\ref{tab:lerf_3dovs} reports the semantic field reconstruction time, highlighting the efficiency of our framework.
LEGaussians, LangSplat and OpenGaussian rely on iterative multi-view optimization, leading to slow convergence. 
In contrast, our method employs a training-free, forward-only pipeline based on a superpoint graph representation.
This design yields up to 30× speedup on LERF-OVS and over 100× on 3DOVS, where fewer input views further amplify the advantage.
While GOI improves efficiency through a feature clustering codebook, it remains within a training-based paradigm and exhibits limited segmentation quality. On the 3DOVS dataset, our method achieves over 30× faster reconstruction compared to GOI, demonstrating superior performance in both accuracy and speed.

\subsection{Ablation Study}
We present ablation results in Table~\ref{tab:ablation}, analyzing the impact of key components in our framework.
Within the Contrastive Gaussian Partitioning module, removing edge reweighting results in less accurate superpoint boundaries, which hinders both hierarchical merging and semantic precision.
Replacing the depth-aware decay with fixed coefficients $\delta_+$ and $\delta_-$ causes the model to over-reliance on reprojection results from regions far away from the camera, reducing boundary reliability.

\begin{table}[h]
  \caption{Evaluation metrics for ablation studies on LERF-OVS~\cite{lerf2023} dataset. ``SR Time'' refers to the Semantic Field Reconstruction Time.}
  \label{tab:ablation}
  \begin{tabular}{ccc}
    \toprule
    Method & \makebox[0.15\linewidth][c]{mIoU}& \makebox[0.15\linewidth][c]{SR Time}\\
    \midrule
    w/o Edge Reweighting                                & 48.50   & 81s\\
    w/o Depth Decay on $\delta_+$ and $\delta_-$        & 51.59    & 84s \\
    Instance Level Only                & 41.70     & \best 35s \\
    w/o Progressive Merging                       & 44.26     & 120s \\
    Ours Full                  & \best 54.94 &  90s  \\
  \bottomrule
\end{tabular}
\end{table}

For the hierarchical semantic representation, using only instance-level SAM masks limits the model’s ability to capture fine-grained semantics. 
Without progressive merging, independently constructing all levels from $\mathcal{S}_0$ undermines hierarchical coherence and leads to over- or under-segmentation. 
These results highlight the importance of both contrastive partitioning and hierarchical representation for achieving accurate 3D semantic understanding.

\section{Conclusion}

We present a novel framework for training-free, open-vocabulary 3D scene understanding by integrating Gaussian Splatting with a hierarchical superpoint graph. 
We begin with contrastive partitioning of Gaussian primitives and progressively merge superpoints under SAM-guided cues to form a multi-level graph structure. An efficient feature reprojection strategy is then used to construct a semantic field aligned with the multi-level superpoint graph.
This training-free pipeline yields semantically consistent and hierarchically structured representations without iterative optimization. Extensive experiments on LERF-OVS, 3DOVS, and ScanNet demonstrate state-of-the-art performance in open-vocabulary segmentation while achieving significant speedup, highlighting the effectiveness of our approach for open-vocabulary 3D scene understanding. 


\bibliographystyle{ACM-Reference-Format}
\bibliography{sample-base}

\clearpage 

\appendix

\section{Evaluation Protocol on ScanNet}

Following the experimental setup of OpenGaussian~\cite{wu2024opengaussian}, we evaluate the performance of 3D semantic segmentation on the ScanNet~\cite{dai2017scannet} dataset. In the following, we detail the specific evaluation protocol on the ScanNet dataset used in OpenGaussian.

During the scene reconstruction stage, we use the raw scanned point clouds provided by the dataset as the initialization for Gaussian Splatting. The densification and position optimization are disabled, keeping all Gaussian primitives fixed to the input point cloud. Only other geometric and appearance attributes are optimized. For semantic field construction, each method utilizes the reconstructed Gaussian scenes in its own way. 

During testing, we classify each GP based on its open-vocabulary features, and assess mIoU and mAcc by comparing the GP predictions with the ground-truth semantic labels of the ScanNet point cloud. This provides an evaluation of 3D point-level semantic understanding.

The predefined categories for classification are derived from the commonly occurring object classes in the ScanNet dataset. The 19 categories (as defined by ScanNet) used for text queries are: wall, floor, cabinet, bed, chair, sofa, table, door, window, bookshelf, picture, counter, desk, curtain, refrigerator, shower curtain, toilet, sink, and bathtub. The 15-category subset excludes picture, refrigerator, shower curtain, and bathtub, while the 10-category subset further excludes cabinet, counter, desk, curtain, and sink.

OpenGaussian evaluates on randomly selected 10 scenes from ScanNet: \textit{scene0000\_00}, \textit{scene0062\_00}, \textit{scene0070\_00}, \textit{scene0097\_00}, \textit{scene0140\_00}, \textit{scene0200\_00}, \textit{scene0347\_00}, \textit{scene0400\_00}, \textit{scene0590\_00}, and \textit{scene0645\_00}. To ensure a fair comparison, we conduct our evaluation on the same set of scenes.

\section{Experimental Details}

\subsection{Interactive Segmentation via Hierarchical Superpoint Graph}
In addition to open-vocabulary querying, our method naturally enables interactive segmentation, owing to its structured and 3D-consistent representation. Unlike methods that rely on per-view 2D supervision, our approach explicitly models semantic entities in 3D space, allowing users to directly interact with the scene at the object or part level.

\begin{figure}[!ht]
    \centering
    \includegraphics[width=1\linewidth]{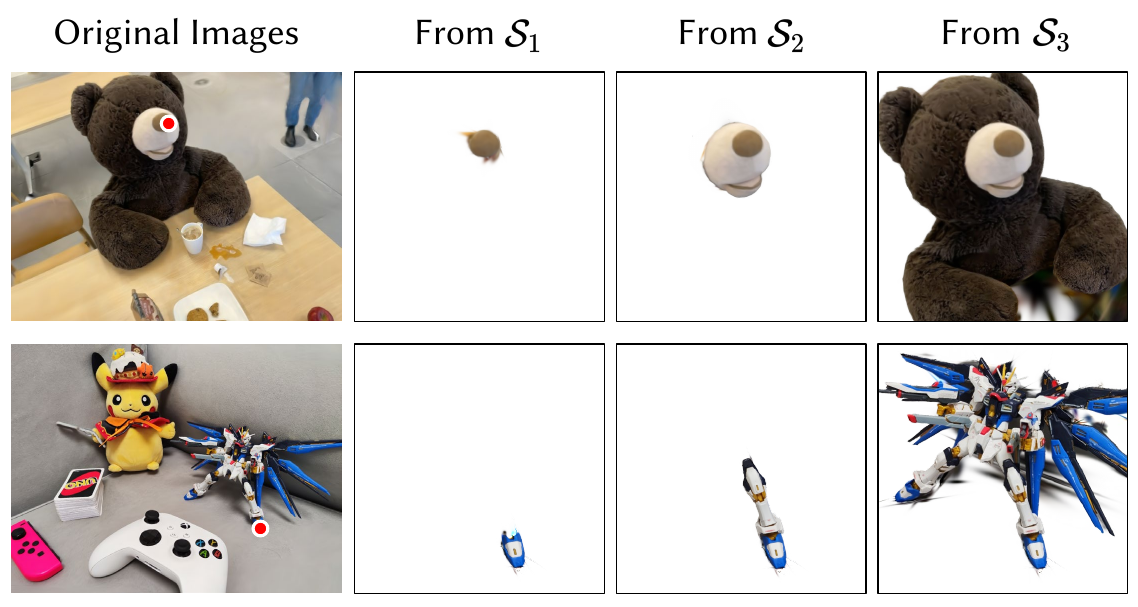}
    
    \caption{
    Interactive Segmentation Results. 
Starting from a point prompt (red circle), our method enables coarse-to-fine object segmentation by traversing the hierarchical superpoint graph—retrieving regions from fine-grained components ($\mathcal S_1$) to complete objects ($\mathcal S_3$). This demonstrates the flexibility of our hierarchical representation for intuitive part-level interaction.
    }
    \label{fig:interactive}
\end{figure}

This capability is further enhanced by our multi-level superpoint graph, which organizes the scene into a semantic hierarchy—from fine-grained components to whole objects. As shown in Figure~\ref{fig:interactive}, a user can interactively select a superpoint corresponding to a queried concept (e.g., nose of the toy bear) and then navigate through the hierarchy to refine or expand the region (e.g., separating wider area of the mouse and the whole toy bear), enabling flexible and semantically coherent scene editing.

Together, these properties make our method well-suited for interactive applications that require accurate and hierarchical 3D scene understanding.

\subsection{Additional Comparisons}

\subsubsection{Qualitative Results.}

\begin{figure*}[!ht]
    \centering
    \includegraphics[width=0.98\textwidth]{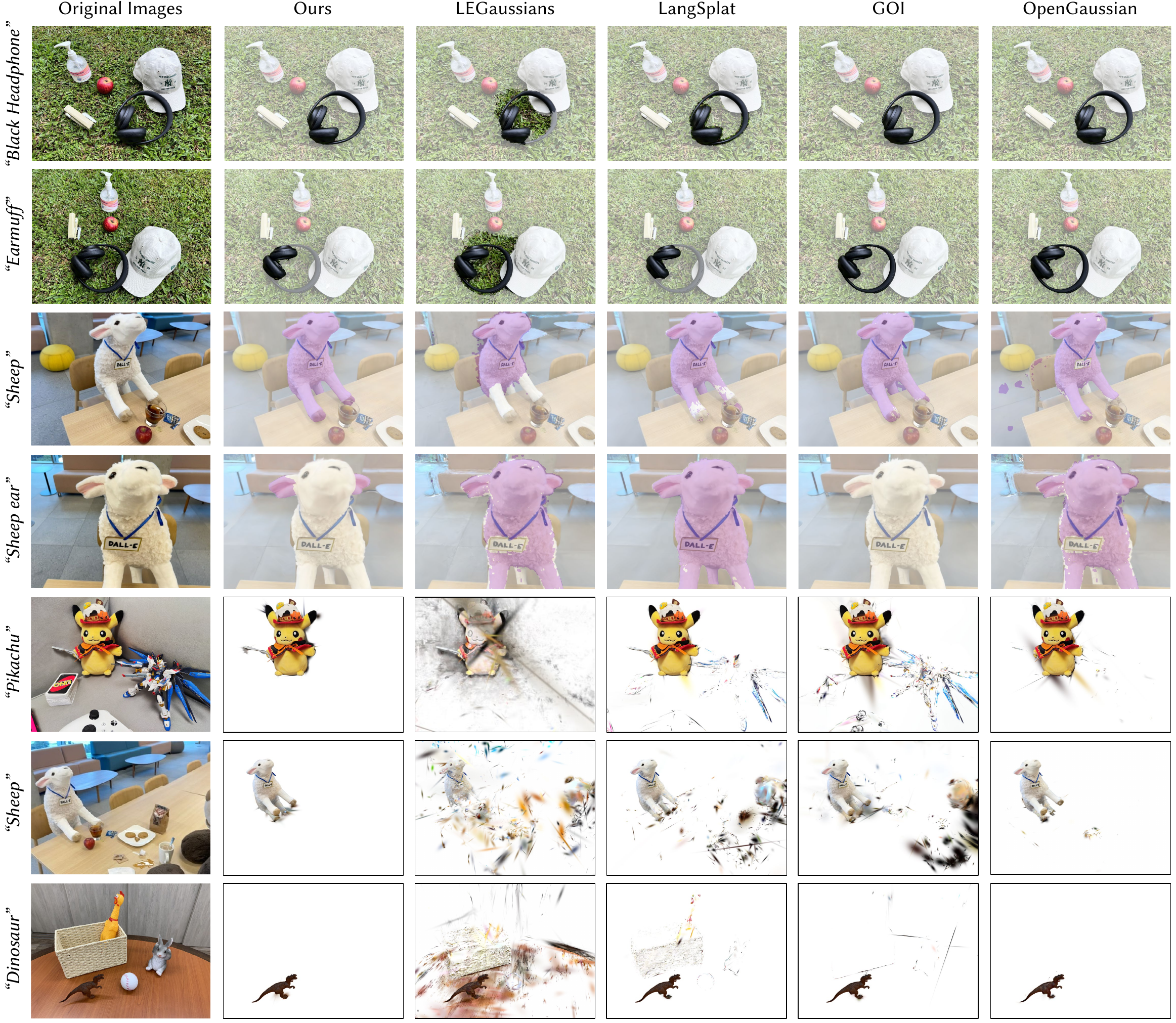}
    
    \caption{
Qualitative comparisons of open-vocabulary on 2D images (top four rows) and 3D Gaussian primitives (bottom three rows). 
We show results from our method alongside LEGaussians~\cite{shi2023legs}, LangSplat~\cite{qin2023langsplat}, GOI~\cite{qu2024goi}, and OpenGaussian~\cite{wu2024opengaussian}. Our approach delivers coherent 3D understanding and effectively supports open-vocabulary querying at both object and part levels. The queried foreground regions are highlighted, and the prompts are shown on the left side of each row. 
    }
    \label{fig:qualitative}
\end{figure*}

Figure~\ref{fig:qualitative} extends the main paper’s comparison by presenting additional qualitative results for both 2D and 3D open-vocabulary segmentation. We compare our method with LEGaussians, LangSplat, GOI, and OpenGaussian, across multiple scenes and query types, including both object-level and part-level descriptions.

Methods that rely on iterative 2D optimization without 3D constraints often produce inconsistent predictions between 2D observations and the underlying 3D scene structure. In contrast, our approach delivers cleaner segmentations with fewer artifacts and more accurate, consistent results. Moreover, it supports multi-level semantic understanding, enabling both coarse object-level and fine-grained part-level segmentation within a unified framework.

\subsubsection{Quantitative Results.}
We evaluate the performance of open-vocabulary segmentation using multiple metrics. In addition to the mIoU results reported in the main paper, Table~\ref{tab:macc} presents the mean class accuracy (mAcc) across all evaluated scenes. This complementary metric further demonstrates the effectiveness and robustness of our framework. 
Our method consistently achieves strong performance across individual scenes and outperforms prior approaches in overall mAcc on both LERF-OVS~\cite{lerf2023} and 3DOVS~\cite{liu20233dovs} datasets, highlighting its capability for accurate and efficient open-vocabulary 3D understanding.

\begin{table*}[t]
  \caption{Additional quantitative comparison of 2D open-vocabulary segmentation on the LERF-OVS~\cite{lerf2023} and 3DOVS~\cite{liu20233dovs} datasets. ``SR Time'' refers to the Semantic Field Reconstruction Time. The best and second-best results are highlighted in \textcolor{tabcapred}{red} and \textcolor{tabcaporange}{orange}.}
  \label{tab:macc}
   {\begin{tabular}{c |cccc c|c |ccccc c|c}
    \toprule
    \multirow{2}{*}{Methods} & \multicolumn{5}{c|}{LERF-OVS mAcc (\%)} & LERF & \multicolumn{6}{c|}{3DOVS mAcc (\%)} & \makebox[0.05\linewidth][c]{3DOVS} \\
     & \makebox[0.055\linewidth][c]{Figurines} & \makebox[0.047\linewidth][c]{Ramen} & \makebox[0.047\linewidth][c]{Teatime} & \makebox[0.047\linewidth][c]{Waldo} & \textbf{Overall} & \makebox[0.05\linewidth][c]{SR Time} 
     & \makebox[0.039\linewidth][c]{Bed} & \makebox[0.039\linewidth][c]{Bench} & \makebox[0.039\linewidth][c]{Lawn} & \makebox[0.039\linewidth][c]{Room} & \makebox[0.039\linewidth][c]{Sofa} &\textbf{Overall}& \makebox[0.05\linewidth][c]{SR Time}\\
    \midrule
    LEGau. \cite{shi2023legs} & 96.65    & 76.86    &92.32  & 79.65  & 86.37 &65min  
    &69.72 &95.60 &96.02 &96.58 &71.59 &85.91  & 75min\\    
    
    LangSplat \cite{lerf2023} & 98.81   &  94.16   & 97.32  & \sbest 95.90 & \sbest 96.55   & 85min    
    &99.17 & \sbest 99.05 & \sbest 99.75 & \best {{99.82}} & \sbest 98.99 & \sbest 99.35 & 90min\\    
    GOI \cite{qu2024goi} & 96.16 & 92.11 & 97.68 & 91.95 & 94.48    & \sbest 15min 
     & \best {{99.78}} & 88.42 & \best {{99.80}} & 99.17 & 94.02 & 96.24  & \sbest 14min \\
     
    OpenGau. \cite{wu2024opengaussian} & \best 99.66  & \sbest 95.11    & \sbest 98.28  & 93.12 & 96.54  & 45min  
    & 75.21 & 88.15 & 73.30 & 72.01 & 94.98 & 80.73   & 55min\\   
    
    Ours & \sbest 99.30 & \best{96.39} & \best {99.13} &\best {97.24}  & \best {98.02} & \best {90s} 
    & \sbest 99.53 & \best {99.58} & {99.72} & \sbest 99.59 & \best {99.36} & \best {99.56}  & \best {25s} \\   

  \bottomrule
\end{tabular}}
\end{table*}

\subsection{Ablation Study: Effect of Gaussian Representation}
To assess the impact of the underlying Gaussian representation, we compare our pipeline built on 2DGS~\cite{Huang2DGS2024} with a variant that uses the original 3DGS~\cite{kerbl20233dgaussian} for scene reconstruction.

While both 2DGS and 3DGS represent scenes using Gaussian primitives, 2DGS adopts surface-aligned disks that better capture object geometry. This alignment preserves object boundaries and reduces ambiguity between nearby objects. As a result, the constructed adjacency graph more accurately captures object-level structure, making it less likely for primitives from different objects to be erroneously grouped into the same superpoint.

In contrast, 3DGS relies on volumetric ellipsoids and often introduces a larger number of floaters, particularly in textureless or blurred regions. These floaters introduce noisy or misleading proximity relations in the graph. This increases the risk of grouping unrelated Gaussians and consequently damages the spatial and semantic coherence of superpoints.

\begin{table}[ht]
  \caption{Evaluation of different Gaussian representations on the LERF-OVS~\cite{lerf2023} dataset.}
  \label{tab:2dgs_3dgs}
  \begin{tabular}{cccc}
    \toprule
    Method & mIoU & mAcc & SR Time\\
    \midrule
    Ours (w/ 2DGS)      & \best {{54.94}}     &\best {{98.02}}     & \best {{90s}} \\
    Ours (w/ 3DGS)          & 42.50    & 96.88   & 102s \\
  \bottomrule
\end{tabular}
\end{table}

Table~\ref{tab:2dgs_3dgs} shows that switching to 3DGS results in a drop in both mIoU and mAcc due to degraded boundary precision and object separation. While the runtime remains comparable, the segmentation quality is noticeably reduced. These findings confirm the advantage of 2DGS as a geometry-aware representation that enables more accurate and spatially consistent semantic field construction.
\clearpage

\end{document}